\documentclass[conference]{IEEEtran}

\usepackage[T1]{fontenc}
\usepackage[utf8]{inputenc}
\usepackage{cite}
\usepackage{url}
\usepackage[inline]{enumitem}
\usepackage{amsmath,amssymb}
\usepackage{balance}

\usepackage{tikz}
\usetikzlibrary{positioning}

\begin{document}

\title{Hardware-Enforced Semantic Coordination for Safety-Critical Real-Time Autonomous Systems}

\author{
\IEEEauthorblockN{Uwe M. Borghoff}
\IEEEauthorblockA{\textit{Department of Computer Science} \\
\textit{University of the Bundeswehr Munich}\\
Neubiberg, Germany \\
uwe.borghoff@unibw.de}
\and
\IEEEauthorblockN{Paolo Bottoni}
\IEEEauthorblockA{\textit{Department of Computer Science} \\
\textit{Sapienza University of Rome}\\
Rome, Italy \\
bottoni@di.uniroma1.it}
\and
\IEEEauthorblockN{Remo Pareschi}
\IEEEauthorblockA{\textit{STAKE Lab} \\
\textit{University of Molise}\\
Campobasso, Italy \\
remo.pareschi@unimol.it}
}

\maketitle
\thispagestyle{plain}
\pagestyle{plain}

\begin{abstract}
Recent advances in agentic AI are producing increasingly complex autonomous systems that integrate large language models, world models, optimization engines, specialized neural architectures, autonomous platforms, and human operators. While much current research focuses on improving reasoning capabilities, safety-critical real-time deployment also requires bounded and verifiable coordination among heterogeneous components operating concurrently under uncertainty. 
Software-mediated coordination presents fundamental limitations in domains where bounded latency, deterministic coordination, and enforceable safety guarantees are essential.

Hence, we propose a hardware-enforced semantic coordination architecture in which selected coordination semantics are implemented directly at the hardware level via field-programmable gate arrays (FPGAs). 
The approach builds on the Topic-Based Communication Space Petri Net (TB--CSPN) framework, which separates semantic reasoning from interaction management.

In this approach, selected TB--CSPN coordination mechanisms are mapped onto FPGA primitives, creating a hardware-native semantic coordination layer. 
Focus is not on acceleration, but on enforcing temporal synchronization, semantic gating, authorization constraints, and bounded coordination behavior directly in hardware. 
Semantic reasoning remains adaptive and software-driven, while embedded coordination semantics become deterministic.
\end{abstract}


\section{Motivation}
The dominant research direction in agentic AI is improving reasoning: better foundation models, better planners, and better integration of symbolic and sub-symbolic components. These advances are real and necessary. 
They are, however, insufficient for safety-critical real-time deployment.
The reason is that real-time autonomous systems do not fail primarily because their reasoning components are weak. They fail because the interactions between reasoning components are unbounded, non-deterministic, and unenforceable. A multi-agent system may misbehave not because any single agent reasons incorrectly, but because messages arrive too late, synchronization conditions are interpreted inconsistently, or authorization steps are bypassed under operational pressure. In real-time autonomous systems, coordination itself is a first-class safety concern.
Current agentic AI systems depend on software-based coordination frameworks built around asynchronous messaging, API-level interaction, and cloud-centric execution models. These mechanisms are well-suited to many non-critical applications, in which occasional delays, retries, or inconsistent response times can be tolerated. In real-time and adversarial environments, however, they face important limitations: scheduling decisions are often non-deterministic, latency can vary unpredictably, and concurrent interactions may give rise to race conditions or ambiguous synchronization behavior. As coordination is typically distributed across software services, message queues, middleware, and remote execution environments, auditing the resulting behavior, or proving enforcement of safety constraints, can be difficult.

The central challenge is therefore not only to make agents more capable, but also to make their interactions more dependable. Software coordination alone provides adaptability, but it does not necessarily provide deterministic timing or non-bypassable safety enforcement. In real-time autonomous systems, coordination itself becomes a first-class safety concern. It is not sufficient for individual agents to reason effectively or generate locally appropriate actions; their interactions must also occur within predictable temporal and logical boundaries. A multi-agent system may fail not because any single reasoning component is defective, but because messages arrive too late, synchronization conditions are interpreted inconsistently, or authorization steps are bypassed under operational pressure.

\begin{figure*}[ht]
\centering
\includegraphics[width=\textwidth]{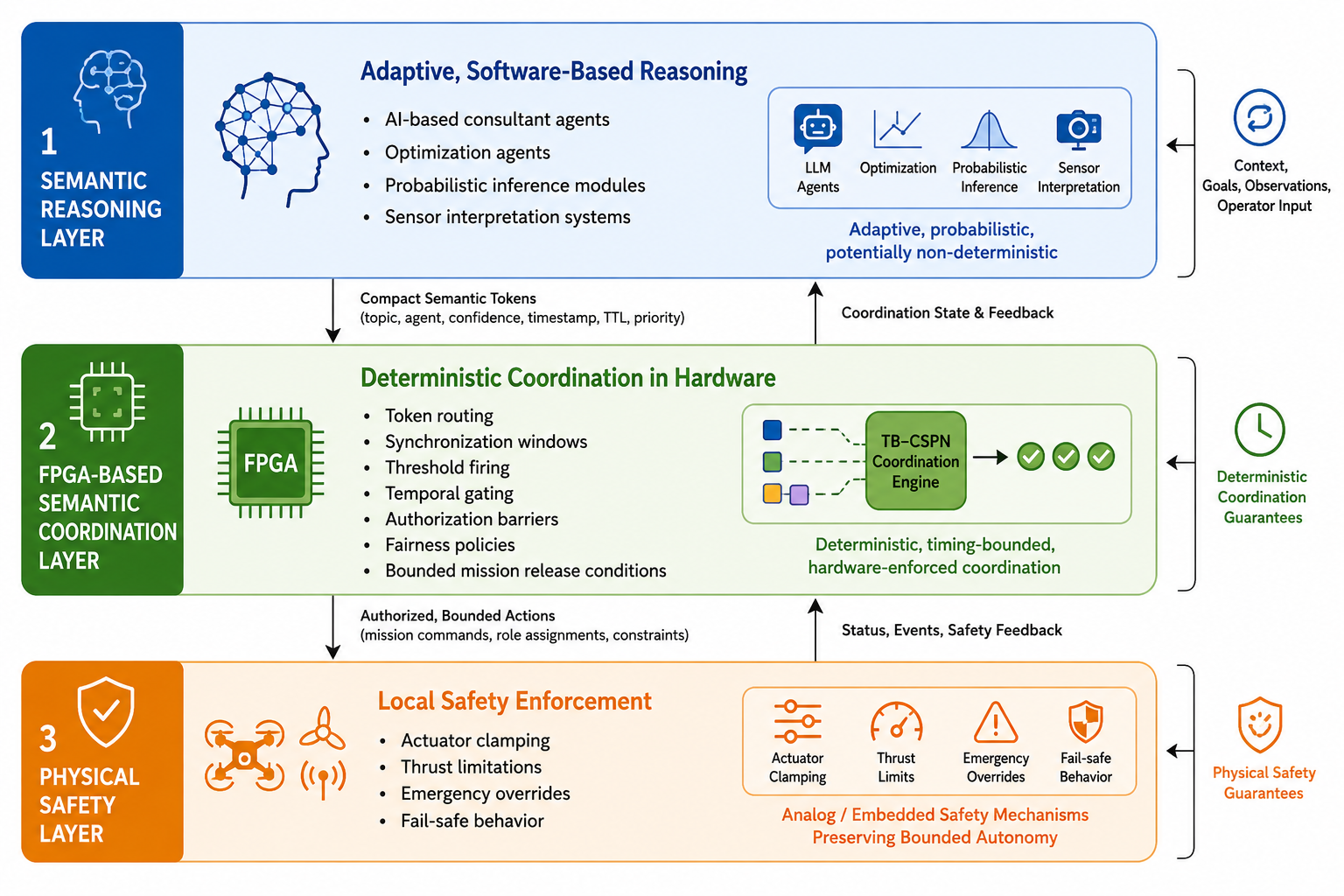}
\caption{Three-layered architecture.}
\label{fig_architecture}
\end{figure*}

One possible approach to addressing this challenge is the TB--CSPN framework, which makes coordination semantics explicit through topic-mediated Petri-net structures. Instead of treating interaction as an informal exchange of messages among software components, TB--CSPN represents coordination as a structured token flow, with synchronization points, bounded decision windows, and authorization conditions defined within the system model. The meaning and operational role of exchanged information therefore become explicit coordination elements within the system model. This provides a clearer separation between what agents reason about and how their interactions are constrained.
Previous work on Guarded Swarms extended this idea through layered autonomy, separating semantic reasoning, coordination, mission formation, and physical safety enforcement. This allows adaptive AI components to remain flexible while lower layers impose stronger operational constraints. The present work continues this trajectory by asking whether the coordination layer itself can be implemented in hardware, turning selected coordination rules into deterministic and enforceable execution properties.

FPGAs are well suited to this role because they combine hardware-level execution with substantial configurability. Unlike fixed-function circuits, they can be reprogrammed to implement different digital logic structures, making them attractive for systems that require both performance and adaptability. They are widely used where low-latency processing, fast digital signal handling, and flexible circuit modification are essential. Consequently, FPGAs are used not only for prototyping but also in deployed systems operating under stringent real-time, resource, and timing constraints.

Their relevance is especially clear in autonomous and defense/security-related applications. 
For example, FPGA-based designs have been used for the real-time onboard classification of military vehicles from drone video streams \cite{VasaviSAF24} and for search-and-rescue operations by UAVs/UGVs \cite{HuangCHYC24}, in which perception must operate with tight latency and limited onboard resources, such as system power \cite{WangDW25}.
Similarly, Zhang \textit{et al.}~\cite{ZhangMK22} demonstrate how FPGA-based CNN acceleration can improve edge inference on resource-constrained drones by reducing off-chip memory access and data movement during real-time processing.
Of course, there are also civilian applications; see, for example, \cite{BellocchiMCPM26} or \cite{GuardunoMartinezCVGRPRO23}.
This illustrates the broader value of FPGA technology for embedded autonomy: computational functions can be placed close to the physical platform and executed with predictable timing.

These properties make FPGAs a natural candidate for implementing the coordination layer proposed in this work. The objective is not to use the FPGA as a general-purpose AI accelerator, but to exploit its deterministic execution model for selected coordination functions. By mapping synchronization, token handling, timing constraints, and authorization logic onto programmable hardware, the system can preserve the adaptability of software-based semantic reasoning while enforcing critical interaction rules at the hardware level.

\section{Architectural Vision}
Figure~\ref{fig_architecture} shows the three layers in the proposed architecture.

\subsection{Semantic Reasoning Layer}
This software layer comprises the system's adaptive reasoning components: AI-based consultant agents, optimization modules, probabilistic inference mechanisms, and sensor interpretation services. These components process mission goals, environmental observations, operator input, and platform-specific constraints to generate recommendations, plans, or candidate actions. 
Rather than deterministic execution guarantees, they provide flexibility and contextual intelligence, relying on, e.g., large language models, world models, statistical estimators, learning-based perception modules, or search and optimization procedures. These methods are inherently adaptive and can produce different outputs depending on context, uncertainty, model state, or input history. This is acceptable and often desirable at the level of semantic reasoning. However, because these components are adaptive, probabilistic, and context-sensitive, a coordination layer is necessary to impose timing, synchronization, authorization, and safety constraints before actions are released to the physical system.

\subsection{FPGA-Based Semantic Coordination Layer}
This layer constitutes the paper's core contribution. It maps selected TB-CSPN coordination mechanisms onto FPGA-executable logic, moving critical interaction rules from software to programmable hardware. The FPGA layer processes compact semantic tokens, routes them based on their topic and agent identifiers, and enforces synchronization conditions before releasing coordinated actions. The FPGA layer implements bounded time windows for mission formation, threshold-based firing rules for collective decisions, and temporal gates that prevent stale or premature tokens from affecting system behavior. The FPGA layer can also support authorization barriers to ensure that human approval or supervisory constraints are met before sensitive transitions occur.

Additionally, the FPGA layer can encode fairness policies and bounded mission-release conditions to mitigate the risk that specific agents, topics, or token classes dominate the coordination process. These functions do not require the FPGA to perform semantic reasoning in the full symbolic or probabilistic sense. Rather, the hardware layer enforces the coordination rules associated with already-interpreted semantic tokens. Thus, adaptive reasoning remains in software, whereas timing-sensitive coordination constraints are enforced with deterministic hardware behavior.

Instead of storing large semantic payloads in hardware, the architecture uses compact semantic tokens composed of indexed metadata fields, which may include a topic identifier, an agent identifier, a confidence level, a timestamp, a time-to-live (TTL) value, and priority flags. The rich semantic content remains in the software layer or external memory while the field-programmable gate array (FPGA) processes only the representation of that content relevant to coordination. This design keeps the hardware logic lightweight, predictable, and suitable for real-time execution. Therefore, the FPGA is not treated as a general symbolic reasoning engine. Rather, it functions as a semantic coordination processor that evaluates token attributes, enforces timing and synchronization rules, and guarantees interaction semantics under deterministic hardware constraints.

\subsection{Physical Safety Layer}
At the lowest level, analog or embedded safety mechanisms provide direct physical protection by operating close to the actuators and enforcing limits independently of higher-level reasoning and coordination. Their functions include actuator clamping, thrust limitation, emergency override, and fail-safe behavior in the event of unreliable communication, computation, or supervision. By placing these safeguards near the physical interface, the system mitigates the risk of software errors, delayed messages, compromised agents, or incorrect mission decisions resulting in unsafe motion.

This preserves the principle of bounded autonomy posited by the Guarded Swarms approach. 
While agents can reason adaptively, coordinate via token-based mechanisms, and pursue mission objectives, local safety enforcement constrains their operational behavior. 
The physical layer acts as a final barrier between digital decision-making and real-world actuation. Without replacing coordination or reasoning, it ensures that autonomy remains operationally bounded, even under degraded or adversarial conditions.

\section{Related Work}
\label{sec:related-work}
The present work builds on our previous research on trusted multi-agent coordination and hardware-supported safety enforcement. 
In~\cite{techdefense/BorghoffBP25}, we combined digital orchestration with hardware-level safeguards for trusted agentic systems. 
This followed the applied UAV monitoring approach in~\cite{carovilla2023integrating}, where blockchain-backed supervision improved accountability but also exposed limitations of centralized coordination, including bottlenecks and single points of failure. 
Related distributed approaches have since employed blockchain and consensus mechanisms to enhance UAV networking and coordination; for example, Zhou \textit{et al.}~\cite{Zhou2024} employ a distributed block consensus algorithm for adaptive UAV communication.

A broader theoretical foundation was developed in~\cite{FRONTIERS/BorghoffBP25}, in which concepts from multi-agent systems~\cite{Wooldridge2002} and Centaurian systems~\cite{centaurian24} were integrated into a communication-space model. 
This line of work ultimately led to the TB--CSPN architecture~\cite{DISCOVER/BorghoffBP25}, which separates semantic reasoning from coordination logic and manages constrained non-determinism using colored Petri nets~\cite{Jensen2009}. 
TB--CSPN supports both autonomous and human-supervised operations while preserving correctness properties such as liveness and bounded execution~\cite{FUTUREINTERNET/BorghoffBP25}. 
A key property of the model is its \emph{sublinear} coordination overhead, as demonstrated through an open-source implementation\footnote{\url{https://github.com/Aribertus/tb-cspn-poc}}. 
Hierarchical token routing and localized concurrency windows reduce communication pressure as swarm size grows, making the architecture suitable for scalable real-time coordination.

The Guarded Swarms framework~\cite{FUTUREINTERNET/BorghoffBP26} extends this trajectory by combining lightweight digital coordination with independent analog safety enforcement for UAV--UGV teams. 
High-level directives from supervisors or AI modules are handled by the token-based coordination layer, while fast analog circuits validate or veto actuator commands. 
This separation preserves flexible mission orchestration while ensuring deterministic physical safeguards, even in the presence of communication loss or cyber compromise. 
A related trust-oriented perspective is presented by Lassfolk \textit{et al.}~\cite{Lassfolk2026}, who propose a certificate-based two-layer trust model and drone-swarm onboarding process for dynamic battlefield scenarios.

A substantial literature addresses trust, resilience, and distributed coordination in UAV and Internet-of-Drones systems~\cite{internet/Hall16,adhoc/BoccadoroSG21,access/LabibBDB21,Rathee2022,vcomm/NairTP25,wu2023q,Iqbal2025,gong2023flight,caballero2024artificial,Westheider2023MultiUAVDRL}, but none addresses the hardware-coordination boundary that is the focus of this work.

Security surveys emphasize the need for robust coordination mechanisms under adversarial conditions. 
Cordill \textit{et al.}~\cite{CordillFX25} review UAV vulnerabilities across hardware, software, and communication layers, including privacy and profiling risks. 
Yu \textit{et al.}~\cite{YuKHE25} classify UAV cyberattacks such as spoofing, jamming, and coordination-level compromise, together with mitigations including encryption, anomaly detection, and fail-safe control. 
At the network layer, Wang \textit{et al.}~\cite{WangYMZXN25} propose a distributed segment-based routing scheme with dynamic \textsc{SRv6} activation for scalable UAV-swarm communication.

Recent defense-oriented systems point in a similar architectural direction, although without the explicit formal token semantics proposed here. 
Helsing's \textsc{HX-2} is an AI-enhanced, software-defined strike UAV with onboard avionics supporting constraints such as geofencing, collision avoidance, and flight-envelope limits\footnote{\url{https://helsing.ai/altra}}. 
\textsc{DARPA}'s \textsc{OFFSET} program explored centralized mission intent combined with distributed swarm execution\footnote{\url{https://www.darpa.mil/research/programs/offensive-swarm-enabled-tactics}}, while the UK \textsc{LANCA} program paired centrally issued maneuver directives with autonomous onboard constraint enforcement\footnote{\url{https://thedefensepost.com/2022/11/03/uk-launches-combat-drone-project/}}. 
On the ground side, \textsc{ARX}'s \textsc{Mithra OS} illustrates the transformation of legacy UGV fleets into interoperable coordinated platforms for contested environments\footnote{\url{https://www.arx-robotics.com/mithra-os}}. 
Together, these developments reflect a broader shift from monolithic autonomy toward layered architectures in which mission-level reasoning, coordination logic, and platform-level safeguards are separated. 
The present paper advances this trend by moving TB--CSPN coordination semantics into an FPGA-backed enforcement layer for real-time agentic systems.

\section{Coordination Rather Than Reasoning}
The proposed architecture adopts a specific stance: that the central engineering challenge of real-time autonomous systems is not improving reasoning but enforcing coordination. Faster inference can improve perception, planning, or decision support, but it does not on its own ensure that agents synchronize correctly, respect authorization constraints, or release actions within predictable temporal bounds.

Accordingly, we propose treating the FPGA layer not as an \textit{AI} accelerator but as a \textit{coordination} accelerator for agentic systems. While the dominant use of FPGAs in agentic and autonomous systems has been to accelerate neural inference or symbolic computation~\cite{VasaviSAF24,ZhangMK22,WangDW25}, our proposal instead enforces coordination semantics directly at the hardware level, in particular by addressing issues related to:

\begin{itemize}
    \item temporal coherence,
    \item synchronization correctness,
    \item interaction constraints, and
    \item bounded coordination latency.
\end{itemize}

The objective is therefore not to replace adaptive semantic reasoning, but to provide a deterministic coordination substrate capable of enforcing timing, synchronization, authorization, and safety constraints independently of the variability of higher-level reasoning components.

\section{Design Challenges}
The proposed architecture raises several research challenges that must be addressed before FPGA-backed TB--CSPN coordination can be realized in practical systems. These challenges concern not only hardware implementation but also the preservation of semantic meaning, temporal correctness, and adaptability under operational constraints.

\subsection{Semantic Token Compactification}

Hardware execution requires compact token representations. TB--CSPN tokens cannot carry arbitrarily large semantic payloads inside the FPGA. Instead, rich semantic content must remain in software components, external memory, or shared knowledge structures, while the hardware layer processes indexed metadata such as topic or agent identifiers, timestamps, confidence values, TTL fields, and priority flags. The challenge is to determine which semantic attributes are essential for coordination and how to encode them without losing the information required for correct interaction management.

\subsection{Ordering and Fairness}

TB--CSPN tokens are semantically differentiated, unlike tokens in many classical Petri-net settings. As a result, token ordering can influence system behavior. A scheduling policy that is efficient from a hardware perspective may still produce undesirable semantic effects, such as delaying high-priority tokens, repeatedly favoring certain agents, or starving specific topics. Fairness policies must therefore be designed together with semantic coherence rules, so that bounded execution does not come at the cost of coordination correctness.

\subsection{Temporal Coordination}

The proposed model also requires explicit temporal semantics. Synchronization windows, TTL expiration, threshold firing, and mission-release conditions depend on time-sensitive token evaluation. These mechanisms differ from classical Petri-net firing rules because transitions may depend not only on token availability, but also on validity intervals, deadlines, and temporal ordering. FPGA implementation must therefore support precise timing while remaining compatible with higher-level coordination semantics.

\subsection{Dynamic Reconfiguration}

A final challenge concerns dynamic reconfiguration in real missions. 
Coordination structures may have to evolve as agents join, leave, fail, or receive new objectives. Supporting such adaptation while preserving hardware-level guarantees remains an open problem and a central direction for future work.

\section{Application Domains}

The proposed architecture is intended for classes of systems in which multiple autonomous or semi-autonomous entities must coordinate under real-time, safety-critical, resource-constrained, or adversarial conditions. In such environments, adaptive reasoning alone is insufficient: interactions among heterogeneous agents must also remain temporally bounded, auditable, and operationally enforceable.

Autonomous drone swarms provide a representative example. Mission execution in these systems depends on timely synchronization, bounded communication, reliable authorization of collective actions, and resilience under degraded communication or adversarial interference. Similar requirements arise in edge-based defense systems, where coordination decisions must often be made near the operational platform under strict latency and safety constraints.

Emergency management and search-and-rescue operations further illustrate the relevance of the approach. In these scenarios, heterogeneous agents combine sensor interpretation, route planning, human supervision, and rapid task allocation while operating in unpredictable or degraded environments. Comparable coordination challenges arise in civilian domains such as precision and smart agriculture, where distributed platforms coordinate monitoring, treatment, and resource allocation across large areas. Critical energy systems, including grid monitoring and infrastructure protection, also require dependable interaction among sensing, control, and response components under temporal and operational constraints.

These examples are representative rather than exhaustive. More generally, the architecture targets systems in which adaptive semantic reasoning must coexist with bounded and enforceable coordination semantics. The broader contribution of the proposed approach therefore lies not in any specific application domain, but in providing a physically grounded coordination substrate for real-time autonomous systems.
\section{Positioning and Scope}
This paper is positioned as a conceptual and architectural contribution rather than an implementation or benchmarking study. It does not present a complete FPGA prototype, hardware synthesis results, or empirical performance measurements. Instead, its purpose is to define the design space for FPGA-backed semantic coordination in real-time agentic systems and to clarify how selected TB--CSPN coordination mechanisms could be mapped onto programmable hardware.

The contribution consists of four main elements. 
First, it proposes a design-oriented architecture separating adaptive semantic reasoning, hardware-enforced coordination, and local physical safety enforcement. 
Second, it outlines an execution model in which compact semantic tokens, temporal gates, synchronization conditions, and authorization barriers are managed by an FPGA-based coordination layer. 
Third, it identifies implementation constraints and trade-offs for token compactification, bounded memory, timing semantics, fairness, dynamic reconfiguration, and the interface between software-reasoning components and hardware logic. Finally, it outlines a research roadmap toward hardware-native semantic coordination, covering formal specification, FPGA realization, verification, and evaluation in representative real-time scenarios.

Accordingly, the paper offers an initial step toward a broader research program. 
Rather than claiming that all coordination logic should be moved to hardware, or that FPGA execution alone solves the challenges of autonomous systems, it argues that selected coordination functions with strong timing and safety requirements may benefit from hardware-level enforcement. The central contribution is therefore to articulate why and how hardware-enforced semantic coordination, grounded in TB--CSPN coordination semantics, could support more deterministic, auditable, and safety-aware autonomous systems.

\section{Discussion and Future Work}

This work proposes to evolve TB--CSPN toward a hardware-backed coordination substrate for real-time autonomous systems. By separating adaptive reasoning from physically enforced coordination, the architecture seeks to combine:
\begin{enumerate*}
    \item flexibility of agentic AI;
    \item deterministic coordination;
    \item bounded autonomy;
    \item enforceable safety constraints.
\end{enumerate*}

The resulting paradigm suggests a shift from software-mediated orchestration toward physically grounded semantic coordination. This may represent an important step toward deployable, verifiable, and safety-aware autonomous systems operating in real-world environments.

An important direction for future work concerns the interaction between hardware-enforced coordination and adaptive mission-level reconfiguration under degraded or adversarial conditions. In the current realization of the Guarded Swarms approach~\cite{FUTUREINTERNET/BorghoffBP26}, when an analog safeguard forces an individual device to abort its local mission segment, the remaining members of the swarm continue executing their assigned tasks. More advanced coordination strategies could allow stress or abort signals to propagate back to the coordination layer, triggering collective reassessment, swarm-wide abort conditions, or dynamic mission redistribution among the remaining agents.

Supporting adaptive coordination while preserving bounded timing, communication constraints, and hardware-level safety guarantees remains an open challenge. Addressing this problem will likely require new coordination semantics and communication mechanisms that operate under strict constraints on latency, bandwidth, and distributed platform geometry.
 
\section*{Acknowledgments}
Figure~\ref{fig_architecture} was generated by \textsc{ChatGPT} based on input from the authors, who reviewed and edited the output and take full responsibility for the content.

\textsc{DeepL Write} and \textsc{ChatGPT} were used to improve the language quality of the paper, which remains an accurate representation of the authors' work and intellectual contributions.

\balance

%


\bibliography{mybibliography}

\begin{thebibliography}{10}
\providecommand{\url}[1]{#1}
\csname url@samestyle\endcsname
\providecommand{\newblock}{\relax}
\providecommand{\bibinfo}[2]{#2}
\providecommand{\BIBentrySTDinterwordspacing}{\spaceskip=0pt\relax}
\providecommand{\BIBentryALTinterwordstretchfactor}{4}
\providecommand{\BIBentryALTinterwordspacing}{\spaceskip=\fontdimen2\font plus
\BIBentryALTinterwordstretchfactor\fontdimen3\font minus
  \fontdimen4\font\relax}
\providecommand{\BIBforeignlanguage}[2]{{%
\expandafter\ifx\csname l@#1\endcsname\relax
\typeout{** WARNING: IEEEtran.bst: No hyphenation pattern has been}%
\typeout{** loaded for the language `#1'. Using the pattern for}%
\typeout{** the default language instead.}%
\else
\language=\csname l@#1\endcsname
\fi
#2}}
\providecommand{\BIBdecl}{\relax}
\BIBdecl

\bibitem{VasaviSAF24}
\BIBentryALTinterwordspacing
S.~Vasavi, D.~S. Sowmya, C.~Aishwarya, and W.~Flores{-}Fuentes, ``{FPGA} based
  military vehicle classification from drone-based video using deep learning,''
  in \emph{9th International Conference on Frontiers of Signal Processing,
  {ICFSP} 2024, Paris, France, September 12-14, 2024}.\hskip 1em plus 0.5em
  minus 0.4em\relax {IEEE}, 2024, pp. 32--37. [Online]. Available:
  \url{https://doi.org/10.1109/ICFSP62546.2024.10785445}
\BIBentrySTDinterwordspacing

\bibitem{HuangCHYC24}
\BIBentryALTinterwordspacing
C.~Huang, Y.~Chen, C.~Hsu, J.~Yang, and C.~Chang, ``Fpga-based {UAV} and {UGV}
  for search and rescue applications: {A} case study,'' \emph{Comput. Electr.
  Eng.}, vol. 119, p. 109491, 2024. [Online]. Available:
  \url{https://doi.org/10.1016/j.compeleceng.2024.109491}
\BIBentrySTDinterwordspacing

\bibitem{WangDW25}
\BIBentryALTinterwordspacing
Z.~X. Wang, C.~Dang, and L.~Wang, ``Personnel search and rescue detector based
  on reconfigurable {FPGA} accelerator: {A} lightweight multi-scale parallel
  drone-mounted detector,'' \emph{{IEEE} Geosci. Remote. Sens. Lett.}, vol.~22,
  pp. 1--5, 2025. [Online]. Available:
  \url{https://doi.org/10.1109/LGRS.2025.3550349}
\BIBentrySTDinterwordspacing

\bibitem{ZhangMK22}
\BIBentryALTinterwordspacing
Z.~Zhang, M.~A.~P. Mahmud, and A.~Z. Kouzani, ``Fitnn: {A} low-resource
  fpga-based {CNN} accelerator for drones,'' \emph{{IEEE} Internet Things J.},
  vol.~9, no.~21, pp. 21\,357--21\,369, 2022. [Online]. Available:
  \url{https://doi.org/10.1109/JIOT.2022.3179016}
\BIBentrySTDinterwordspacing

\bibitem{BellocchiMCPM26}
\BIBentryALTinterwordspacing
G.~Bellocchi, D.~Madro{\~{n}}al, A.~Capotondi, F.~Palumbo, and A.~Marongiu,
  ``An fpga-based accelerator design methodology for smart uavs in precision
  agriculture: {A} case study,'' \emph{J. Syst. Archit.}, vol. 170, p. 103592,
  2026. [Online]. Available: \url{https://doi.org/10.1016/j.sysarc.2025.103592}
\BIBentrySTDinterwordspacing

\bibitem{GuardunoMartinezCVGRPRO23}
\BIBentryALTinterwordspacing
E.~Guardu{\~{n}}o{-}Martinez \emph{et~al.}, ``An {FPGA} smart camera
  implementation of segmentation models for drone wildfire imagery,'' in
  \emph{Advances in Computational Intelligence - 22nd Mexican International
  Conference on Artificial Intelligence, {MICAI} 2023, Yucat{\'{a}}n, Mexico,
  November 13-18, 2023}, ser. Lecture Notes in Computer Science, H.~Calvo,
  L.~Mart{\'{\i}}nez{-}Villase{\~{n}}or, and H.~E. Ponce, Eds.\hskip 1em plus
  0.5em minus 0.4em\relax Springer, 2023, pp. 213--226. [Online]. Available:
  \url{https://doi.org/10.1007/978-3-031-47765-2\_16}
\BIBentrySTDinterwordspacing

\bibitem{techdefense/BorghoffBP25}
U.~M. Borghoff, P.~Bottoni, and R.~Pareschi, ``Guarded swarms: Hybrid
  digital–analog coordination for {AI}–robot systems,'' in
  \emph{Proceedings of the {IEEE} International Workshop on Technologies for
  Defense and Security ({TechDefense}), Rome, Italy, November 5--7, 2025},
  2025, pp. 289--294.

\bibitem{carovilla2023integrating}
\BIBentryALTinterwordspacing
A.~Carovilla, R.~Pareschi, and F.~Salzano, ``Integrating blockchain for
  enhanced coordination and security in semi-centralized robotic swarms,'' in
  \emph{Proceedings of the {IEEE} International Workshop on Technologies for
  Defense and Security ({TechDefense}), Rome, Italy, November 20--22, 2023},
  2023, pp. 95--99. [Online]. Available:
  \url{https://doi.org/10.1109/TechDefense59795.2023.10380842}
\BIBentrySTDinterwordspacing

\bibitem{Zhou2024}
\BIBentryALTinterwordspacing
X.~Zhou, L.~Yang, L.~MA, and H.~He, ``Towards secure and resilient unmanned
  aerial vehicles swarm network based on blockchain,'' \emph{IET Blockchain},
  vol.~4, no.~S1, pp. 483--493, 2024. [Online]. Available:
  \url{https://doi.org/10.1049/blc2.12050}
\BIBentrySTDinterwordspacing

\bibitem{FRONTIERS/BorghoffBP25}
\BIBentryALTinterwordspacing
U.~M. Borghoff, P.~Bottoni, and R.~Pareschi, ``Human-artificial interaction in
  the age of agentic {AI}: {A} system-theoretical approach,'' \emph{Frontiers
  in Human Dynamics}, vol. 7: 1579166, pp. 1--16, 2025. [Online]. Available:
  \url{https://doi.org/10.3389/fhumd.2025.1579166}
\BIBentrySTDinterwordspacing

\bibitem{Wooldridge2002}
M.~J. Wooldridge, \emph{Introduction to Multiagent Systems}.\hskip 1em plus
  0.5em minus 0.4em\relax Wiley: West Sussex, England, 2002.

\bibitem{centaurian24}
\BIBentryALTinterwordspacing
R.~Pareschi, ``Beyond human and machine: An architecture and methodology
  guideline for {C}entaurian design,'' \emph{Sci}, vol. 6(4): 71, 2024.
  [Online]. Available: \url{https://doi.org/10.3390/sci6040071}
\BIBentrySTDinterwordspacing

\bibitem{DISCOVER/BorghoffBP25}
\BIBentryALTinterwordspacing
U.~M. Borghoff, P.~Bottoni, and R.~Pareschi, ``An organizational theory for
  multi-agent interactions integrating human agents, {LLMs}, and specialized
  {AI},'' \emph{Discover Computing}, vol. 28(1): 138, pp. 1--35, 2025.
  [Online]. Available: \url{https://doi.org/10.1007/s10791-025-09667-2}
\BIBentrySTDinterwordspacing

\bibitem{Jensen2009}
\BIBentryALTinterwordspacing
K.~Jensen and L.~M. Kristensen, \emph{Coloured {Petri} Nets - Modelling and
  Validation of Concurrent Systems}.\hskip 1em plus 0.5em minus 0.4em\relax
  Springer, 2009. [Online]. Available: \url{https://doi.org/10.1007/b95112}
\BIBentrySTDinterwordspacing

\bibitem{FUTUREINTERNET/BorghoffBP25}
\BIBentryALTinterwordspacing
U.~M. Borghoff, P.~Bottoni, and R.~Pareschi, ``Beyond prompt chaining: The
  {TB-CSPN} architecture for agentic {AI},'' \emph{Future Internet}, vol.
  17(8): 363, pp. 1--27, 2025. [Online]. Available:
  \url{https://doi.org/10.3390/fi17080363}
\BIBentrySTDinterwordspacing

\bibitem{FUTUREINTERNET/BorghoffBP26}
\BIBentryALTinterwordspacing
------, ``Guarded swarms: Building trusted autonomy through digital
  intelligence and physical safeguards,'' \emph{Future Internet}, vol. 18(1):
  64, pp. 1--30, 2026. [Online]. Available:
  \url{https://doi.org/10.3390/fi18010064}
\BIBentrySTDinterwordspacing

\bibitem{Lassfolk2026}
\BIBentryALTinterwordspacing
C.~Lassfolk and H.~Kari, ``A trust management concept for secure onboarding of
  military coalition drone swarms,'' in \emph{Cyber Security: Policy and
  Technology}, M.~Lehto and P.~Neittaanm{\"a}ki, Eds.\hskip 1em plus 0.5em
  minus 0.4em\relax Springer, Cham, 2026, pp. 193--221. [Online]. Available:
  \url{https://doi.org/10.1007/978-3-032-08890-1_9}
\BIBentrySTDinterwordspacing

\bibitem{internet/Hall16}
\BIBentryALTinterwordspacing
R.~J. Hall, ``An internet of drones,'' \emph{{IEEE} Internet Comput.}, vol.~20,
  no.~3, pp. 68--73, 2016. [Online]. Available:
  \url{https://doi.org/10.1109/MIC.2016.59}
\BIBentrySTDinterwordspacing

\bibitem{adhoc/BoccadoroSG21}
\BIBentryALTinterwordspacing
P.~Boccadoro, D.~Striccoli, and L.~A. Grieco, ``An extensive survey on the
  internet of drones,'' \emph{Ad Hoc Networks}, vol. 122, p. 102600, 2021.
  [Online]. Available: \url{https://doi.org/10.1016/J.ADHOC.2021.102600}
\BIBentrySTDinterwordspacing

\bibitem{access/LabibBDB21}
\BIBentryALTinterwordspacing
N.~S. Labib, M.~R. Brust, G.~Danoy, and P.~Bouvry, ``The rise of drones in
  internet of things: {A} survey on the evolution, prospects and challenges of
  unmanned aerial vehicles,'' \emph{{IEEE} Access}, vol.~9, pp.
  115\,466--115\,487, 2021. [Online]. Available:
  \url{https://doi.org/10.1109/ACCESS.2021.3104963}
\BIBentrySTDinterwordspacing

\bibitem{Rathee2022}
\BIBentryALTinterwordspacing
G.~Rathee, A.~Kumar, C.~A. Kerrache, and R.~Iqbal, ``A trust-based mechanism
  for drones in smart cities,'' \emph{IET Smart Cities}, vol.~4, no.~4, pp.
  255--264, 2022. [Online]. Available: \url{https://doi.org/10.1049/smc2.12039}
\BIBentrySTDinterwordspacing

\bibitem{vcomm/NairTP25}
\BIBentryALTinterwordspacing
A.~S. Nair, S.~M. Thampi, and J.~V.~V. P., ``{SoCoMNNet}: {A} sociocognitive
  and memristive neural network-based context-aware {GPS} spoofing detection
  and mitigation in the internet of drones,'' \emph{Veh. Commun.}, vol.~56, p.
  100980, 2025. [Online]. Available:
  \url{https://doi.org/10.1016/J.VEHCOM.2025.100980}
\BIBentrySTDinterwordspacing

\bibitem{wu2023q}
\BIBentryALTinterwordspacing
M.~Wu, Z.~Zhu, Y.~Xia, Z.~Yan, X.~Zhu, and N.~Ye, ``A {Q}-learning-based
  two-layer cooperative intrusion detection for internet of drones system,''
  \emph{Drones}, vol.~7, no.~8, p. 502, 2023. [Online]. Available:
  \url{https://doi.org/10.3390/drones7080502}
\BIBentrySTDinterwordspacing

\bibitem{Iqbal2025}
\BIBentryALTinterwordspacing
D.~Iqbal, H.~Bangui, and B.~Rossi, ``Trial by twin: Behavior-predictive trust
  in autonomous drone swarms,'' in \emph{Cooperative Information Systems - 31st
  International Conference, CoopIS 2025, Marbella, Spain, October 20-22, 2025,
  Proceedings}, ser. Lecture Notes in Computer Science, C.~Cappiello,
  O.~Hartig, M.~Sellami, and A.~Ouni, Eds., vol. 15535.\hskip 1em plus 0.5em
  minus 0.4em\relax Springer, 2025, pp. 673--683. [Online]. Available:
  \url{https://doi.org/10.1007/978-3-032-15538-2\_44}
\BIBentrySTDinterwordspacing

\bibitem{gong2023flight}
\BIBentryALTinterwordspacing
Y.~Gong and X.~Liu, ``Flight state recognition for {UAV} optical flow velocity
  measurement,'' \emph{Journal of Physics: Conference Series}, vol. 2561,
  no.~1, p. 012025, 2023. [Online]. Available:
  \url{https://doi.org/10.1088/1742-6596/2561/1/012025}
\BIBentrySTDinterwordspacing

\bibitem{caballero2024artificial}
\BIBentryALTinterwordspacing
D.~Caballero-Martin, J.~M. Lopez-Guede, J.~Estevez, and M.~Gra{\~n}a,
  ``Artificial intelligence applied to drone control: {A} state of the art,''
  \emph{Drones}, vol.~8, no.~7, p. 296, 2024. [Online]. Available:
  \url{https://doi.org/10.3390/drones8070296}
\BIBentrySTDinterwordspacing

\bibitem{Westheider2023MultiUAVDRL}
\BIBentryALTinterwordspacing
J.~Westheider, J.~R{\"u}ckin, and M.~Popovi{\'c}, ``{Multi-UAV} adaptive path
  planning using deep reinforcement learning,'' in \emph{IEEE/RSJ International
  Conference on Intelligent Robots and Systems (IROS)}, 2023, pp. 649--656.
  [Online]. Available: \url{https://doi.org/10.1109/IROS55552.2023.10342516}
\BIBentrySTDinterwordspacing

\bibitem{CordillFX25}
\BIBentryALTinterwordspacing
B.~Cordill, D.~Fang, and S.~Xu, ``A comprehensive survey of security and
  privacy in {UAV} systems,'' \emph{{IEEE} Access}, vol.~13, 2025. [Online].
  Available: \url{https://doi.org/10.1109/ACCESS.2025.3583985}
\BIBentrySTDinterwordspacing

\bibitem{YuKHE25}
\BIBentryALTinterwordspacing
A.~Yu, I.~Kolotylo, H.~A. Hashim, and A.~E.~E. Eltoukhy, ``Electronic warfare
  cyberattacks, countermeasures, and modern defensive strategies of {UAV}
  avionics: {A} survey,'' \emph{{IEEE} Access}, vol.~13, 2025. [Online].
  Available: \url{https://doi.org/10.1109/ACCESS.2025.3561068}
\BIBentrySTDinterwordspacing

\bibitem{WangYMZXN25}
\BIBentryALTinterwordspacing
Z.~Wang, H.~Yao, T.~Mai, R.~Zhang, Z.~Xiong, and D.~Niyato, ``Toward
  intelligent distributed segment-based routing in {6G}-era ultra-large-scale
  {UAV} swarm networks,'' \emph{{IEEE} Commun. Mag.}, vol.~63, no.~6, pp.
  58--64, 2025. [Online]. Available:
  \url{https://doi.org/10.1109/MCOM.001.2400583}
\BIBentrySTDinterwordspacing

\end{thebibliography}

\end{document}